\documentclass[11pt]{article}
\usepackage{amsfonts} 
\usepackage[preprint]{acl}

\usepackage{times}
\usepackage{latexsym}

\usepackage[T1]{fontenc}

\usepackage{float}  
\usepackage[utf8]{inputenc}
\usepackage{amsmath} 
\usepackage{comment} 
\usepackage{booktabs} 
\usepackage{microtype}
\usepackage{tabularx}

\usepackage{inconsolata}

\usepackage{graphicx}
\usepackage{color}
\usepackage{multirow}
\usepackage{pdfcomment}

\usepackage[most]{tcolorbox}
\usepackage{tabularx}
\usepackage{booktabs} 
\usepackage{blindtext} 
\usepackage{float}

\usepackage{caption}
\title{
Integrating the Analytic Hierarchy Process with Large Language Models for Transparent Multi-Criteria Decision-Making %
}

{\footnotesize 
    \author{Han Zhiguang$^{1}$, Farah Benamara$^{2,3}$ and Pascale Zaraté$^{4}$ \\
  $^{1}$ CNRS@CREATE, Singapore, Singapore \\
 $^{2}$ IRIT, Université de Toulouse, Toulouse, France \\
  $^{3}$ IPAL-CNRS-NUS-A*STAR, Singapore \\  
 $^{4}$ IRIT, Université Toulouse Capitole, Toulouse, France
    \\
    \textbf{Corresponding author}: farah.benamara@irit.fr} }

\begin{document}
\maketitle


\begin{abstract}
LLMs are increasingly employed in a wide range of decision-making tasks. However, the opacity of their internal reasoning makes it difficult to validate or interpret their outputs, and the need for interpretability becomes especially critical in high-stakes settings. This study examines the decision-making capabilities of LLMs through the Analytic Hierarchy Process (AHP), a classical and widely used multicriteria decision-making framework. We construct a new annotated benchmark based on AHP and propose the first end-to-end approach that enables LLMs to perform the complete AHP workflow. 
Experiments in real-world decision problems in the legal and higher-education ranking domains show that our method significantly improves alignment with expert judgments. 

\end{abstract}

%
\section{Introduction}

Multi-Criteria Decision Making (MCDM) provides systematic approaches for addressing complex problems that involve multiple, often conflicting objectives. Such problems are ubiquitous: policymakers must balance fairness against efficiency, healthcare professionals weigh treatment effectiveness against cost, and legal experts reconcile principles of justice with practical enforceability. Over decades, a rich set of MCDM techniques have been developed, including TOPSIS \cite{papathanasiou2018topsis}, PROMETHEE \cite{brans2005promethee}, and ELECTRE \cite{govindan2016electre}. 
Among these, we focus here on the Analytic Hierarchy Process (AHP)~\cite{saaty1980ahp, forman2001ahp, munier2021uses} (see \citet{simon1960new} and \citet{ opricovic2007extended} for an overview). 

\begin{figure}[t!]
    \centering
    \includegraphics[scale=0.25]{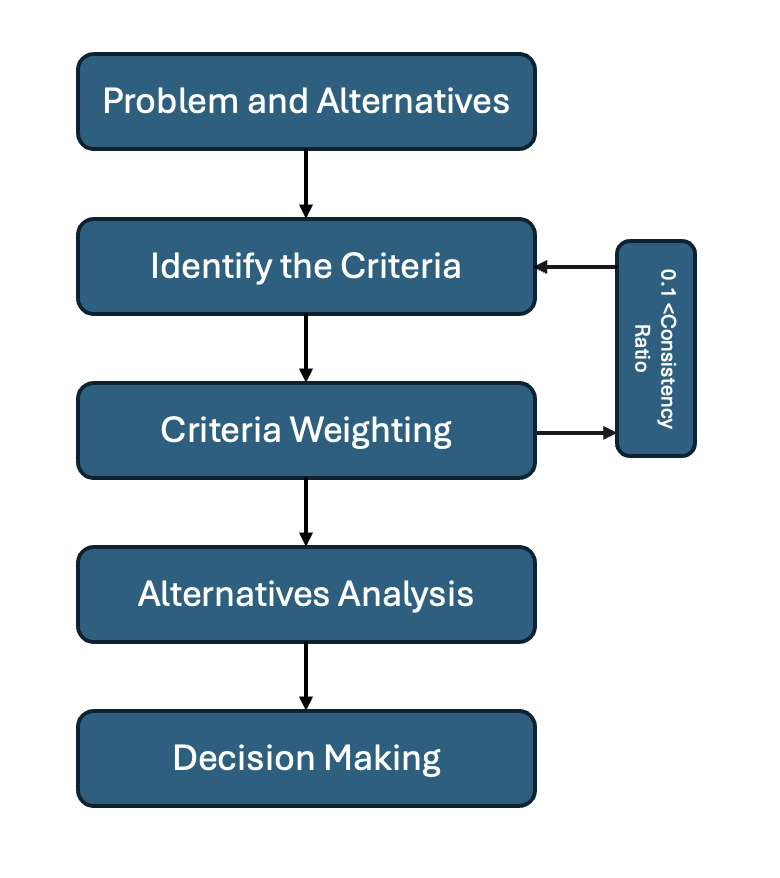}
    \caption{Flowchart of MCDM and AHP. 
    See Appendix \ref{appendix:ahp} for theoretical foundations.  
    }
    \label{fig:ahp}

\end{figure}

AHP addresses decision complexity by decomposing a central goal into a hierarchy of criteria and alternatives (cf. 
Figure~\ref{fig:ahp}). Decision makers conduct pairwise comparisons to evaluate the relative importance of criteria and preferences among alternatives. From these judgments, AHP derives explicit priority weights, checks their consistency (i.e. ratio $<0.1$), and synthesizes them into a final decision. This structured workflow makes AHP
 one of the most influential and widely applied methods in domains ranging from 
 public policy \citep{ho2008integrated}, software selection \citep{labib2011decision}, and healthcare \citep{badri2001ahp}.
Its core strength lies in providing a transparent and auditable reasoning process. 

Recent research has sought to improve the scalability and robustness of AHP, including  fuzzy AHP \citep{soltanpanah2018application, khosla2021fuzzy} which extends the traditional framework to handle uncertainty in expert judgments, and trace-based approaches  leveraging behavioral data to automate criteria weighting \citep{tracebased2021}. However, these methods often depend on predefined knowledge bases or 
manual inputs, which limits their flexibility in real scenarios. 
In this paper, we explore to what extent LLM can be an alternative  to automate the entire AHP reasoning process.



While the use of LLM in decision-support tools has been largely explored in the literature \cite{eigner2024determinants,sha2023languagempc,LLM-Invest,sun2025llm_madm}, AHP-based  approaches received less attention. 
Among the few attempts, 
existing work share three main limitations:  (1) 
AHP prompting adopt simple prompting strategies, yet their findings remain inconclusive, some report strong performance from LLMs \cite{xu2024toward,svoboda2024enhancing}, while others highlight inconsistent or unreliable results \cite{canchatgpt2023,stelmach2025evaluating}. This disparity suggests that conventional static prompting alone may fail to capture the complex reasoning AHP requires.  (2) Comparisons between LLM  and human experts are often constrained by the absence of standardized, publicly available datasets. Furthermore, many studies omit systematic consistency verification, making it difficult to evaluate the reliability and reproducibility of their findings.   (3) Current evaluation practices also tend to overlook the intermediate reasoning steps within the AHP process. Except for \citet{park2025enhancing}, who manually analyzed these steps, most prior studies focus solely on final outcomes, without examining how decisions are formed throughout the pipeline. Our work addresses these gaps by proposing: \\
--  \textbf{\textsc{Legal-AHP},  \textit{a new  expert-annotated benchmark  following AHP}}. We target 
real-world  reasoning in the legal domain, providing a basis for future research in the field 
(cf. Section \ref{sec:dataset}).\footnote{The dataset is available in the Supplementary Material.} \\
\indent -- \textbf{\textit{The first end-to-end 
approach that operationalizes the \emph{entire} AHP workflow with LLMs.}} We propose a set of strategies 
enabling 
interpretable decision-making through  explicit AHP instructions, 
a single-agent AHP framework and 
a multi-agent architecture 
(cf. Section \ref{sec:exp}).\\ 
-- \textbf{\textit{A quantitative and qualitative evaluation of these strategies}} exploring 
both the internal stages of the pipeline and the final decisions, ensuring methodological robustness and transparency throughout (cf. Section \ref{sec:res}).  Our findings 
demonstrate improved alignment with expert judgment and more trustworthy explanations. 
\\
-- \textit{\textbf{An evaluation of the portability of our approach beyond the legal domain}}, targeting ranking tasks. Our results show that LLM-based AHP framework is applicable to various decision problems. 

\section{Related Work}
\subsection{LLMs as a Tool  for Decision-Making}

LLMs are increasingly explored as decision-support tools across domains. \citet{eigner2024determinants} review key determinants of LLM-assisted decision-making, such as transparency, task complexity, and user trust. Beyond support, LLMs are also studied as autonomous decision-makers: for example, \emph{LanguageMPC} \citep{sha2023languagempc} uses an LLM to reason about driving scenarios and translate them into control actions via model predictive control. At a broader scale, \citet{sun2025llm_madm} survey LLM-based multi-agent decision-making, highlighting challenges of coordination, communication, and human oversight. Complementing these efforts, \citet{lu2024strux} propose \emph{STRUX}, a framework that improves transparency by extracting favorable and adverse evidence, weighting them, and generating structured explanations in high-stakes tasks like financial forecasting. 

However, their application to MCDM remains limited: while capable of producing plausible outputs, LLMs lack the structured logic of established methods like AHP, and their reasoning processes are opaque and difficult to verify. As a result, their outputs often cannot meet the reliability and interpretability demands of multi-criteria settings.

Explainable AI (XAI) methods have attempted to mitigate this “black-box” problem, for example through saliency maps and post-hoc rationalizations~\citep{ribeiro2016lime, shapley1953value}. Yet these methods typically provide feature-level explanations and fail to capture how models weigh multiple criteria to reach a final outcome. In MCDM contexts, where process transparency and traceability are paramount, such explanations remain insufficient, especially in high-stakes environments where errors can have serious consequences.

\subsection{LLMs-based  AHP Decision-Making}

Within this broader trend, researchers have explored integrating LLMs with structured decision-science methods such as  AHP. \citet{xu2024toward} introduced “AHP-Powered LLM Reasoning,” where the AHP hierarchy guides prompt construction to make evaluations of open-ended student responses more systematic. Results show that multiple criteria performs better compared to pairwise comparison without criteria.

\citet{canchatgpt2023} tested ChatGPT as a multi-criteria decision-maker and found that although outputs appeared plausible, they often violated AHP's consistency requirements.  Similarly, \citet{svoboda2024enhancing} integrated GPT-4 with  AHP for cybersecurity decision support, employing GPT-4 based “virtual experts” to perform pairwise comparisons and aggregate judgments, albeit only at partial workflow levels.

Most recently, \citet{park2025enhancing} enhanced AHP modelling under uncertainty by fine-tuning LLMs on domain-specific documents to generate complete hierarchies and comparison matrices, showing alignment with expert judgments even under incomplete specifications.

In this paper, we continue these efforts by exploring this time LLMs reasoning capabilities in mimicking the full AHP pipeline while providing to the community the first  benchmark dataset manually annotated following AHP in the legal domain.


\section{The \textsc{Legal-AHP} Dataset }\label{sec:dataset}
There is currently a lack of datasets specifically designed to evaluate MCDM problems, leading most existing studies to rely on multi-attribute datasets such as \citet{cheng2024adapting}.  For AHP, as no prior free  dataset exists,  we propose to leverage on  available question answering (QA) benchmarks that we augment with expert  annotations following an AHP-based annotation scheme enabling a direct comparison between human judgments and the outputs produced by LLMs. To this end, we rely on 
the \textsc{LegalBench} benchmark \cite{holzenberger2023legalbench} which provides a diverse set of carefully curated   problems that closely reflect real-world legal reasoning. Its domain specificity and structured tasks make it a natural choice for evaluating complex decision-making in legal contexts.

\subsection{Dataset Selection}
\textsc{LegalBench} is a collaboratively constructed benchmark, 
comprising 162 subsets or tasks that span six distinct categories of legal reasoning in English. Following a thorough inspection, we sample five subsets, each containing  series of QA for a total of 525 pairs:  45 from \textit{Abercrombie} (9 series of 5 QA), 135 \textit{Judicial\_Ethics} (15 series of 9 QA), 90 \textit{Common\_Law}  (15 series of 6 QA), 120 \textit{Decision\_Section}  (15 series of 8 QA), and 135 \textit{Privacy\_Policy}  (15 series of 9 QA). These subsets are selected for two main reasons: (1) They cover diverse areas of law, ensuring a variety of legal contexts; and (2) They feature distinct decision-making formats (e.g., binary Yes/No judgments and multiple-choice style questions), allowing us to assess the decision-making performance across different task structures.  A brief description of each subset is provided in Appendix~\ref{appendix:subsets}.

\subsection{Annotation Procedure}
The selected subsets were reframed as decision-making problems, with multiple-choice options serving as the alternatives. For each subset, ten independent groups of law Master students with relevant expertise were invited to provide annotations. Each group, composed of 2 members, was randomly given two subsets to enable stronger cross-validation. 

We designed an AHP annotation guidelines where for each QA pair,  annotators defined a list of  3 to 7 criteria  along with  their definitions, constructed pairwise comparison matrices, and computed the weights for each criterion based on these matrices. See Appendix~\ref{appendix:ahp_details} for a detailed description of the guidelines together with a complete example.

A total of 525 questions were annotated in two steps. First a training phase where for each subset,  groups were  trained on the first questions of each series for a total of 450 QA (see Appendix \ref{training} for detail). Given the complexity of the task, the training step was very deep guided by an expert in MCDM.
Then students  were asked to annotate the last questions of each subset (series) separately, which corresponds to a  total of 75 pairs. 
Within a group, annotations are made by consensus among its members. 

\begin{table*}[h]
\centering
\small
\renewcommand{\arraystretch}{1.2}
\begin{tabular}{|p{1.8cm}|p{2.0cm}|p{2.0cm}|p{2.2cm}|p{2.0cm}|p{2.2cm}|}
\hline
\textbf{Criteria} & \textbf{Abercrombie} & \textbf{Common\_Law} & \textbf{Decision\_Section} & \textbf{Judicial\_Ethics} & \textbf{Privacy\_Policy} \\ \hline
\textbf{Criterion 1} 
& Type product/ service 
& Tangibility 
& References to passed cases 
& Impartiality 
& Topic precision \\ \hline
\textbf{Criterion 2} 
& Link product mark 
& Services 
& Affirmation 
& Compliance with law 
& Lexical scope consistency \\ \hline
\textbf{Criterion 3} 
& Originality 
& Real estate 
& Law references 
& Conflict of interest 
& --- \\ \hline
\end{tabular}
\caption{Gold criteria per subset in \textsc{Legal-AHP}.}
\label{tab:criteria_groups}
\end{table*}

\subsection{Qualitative Evaluation}
\paragraph{Consistency check.}
We conducted a consistency check to evaluate the reliability of the criteria and weights provided by the annotators  (see Appendix~\ref{appendix:ahp}, Consistency section). 
A group is retained only when the consistency ratio associated to its annotations satisfied $\text{CR} < 0.1$ or the list of the annotated criteria is greater than two (since pairwise comparisons and the corresponding CR could not be computed). 
Among the 20 annotations from the groups, 
12 were removed due to invalid or inconsistent criteria 
ensuring that only coherent and reliable annotations were included in the analysis. 
Table~\ref{tab:consistency_ratios_group} reports the CR values for the remaining groups across  subsets with $\text{CR} < 0.1$, which corresponds to a total of 60 reliable QA pairs  in the final \texttt{Legal-AHP} benchmark.

\begin{table}[h]
\centering
\small
\setlength{\tabcolsep}{3pt}
\begin{tabular}{@{}lll@{}}
\toprule
\textbf{Subset} & \textbf{Groups} & \textbf{CR$< 0.1$} \\
\midrule
\multirow{3}{*}{Abercrombie} & Group\_1 & 0.0439 \\
 & Group\_2 & 0.0432 \\
 & Group\_3 & 0.0473 \\
\midrule
\multirow{2}{*}{Judicial\_Ethics} & Group\_4 & 0.0348 \\
& Group\_6 & 0.0465 \\
\midrule
\multirow{2}{*}{Common\_Law} & Group\_2 & 0.0332 \\
 & Group\_3 & 0.0334 \\
\midrule
Decision\_Section & Group\_4 & 0.0692 \\
\bottomrule
\end{tabular}
\caption{Consistency Ratios (CR) of annotation groups across subsets.}
\label{tab:consistency_ratios_group} %
\end{table}

\paragraph{Agreements among criteria.} In addition, we manually identified the criteria that were common across groups under the same subset to arrive at a gold list. 
To this end, we rely on the 
definitions provided during annotations and performed a semantic analysis by grouping similar criteria.  
When the definitions were not aligned, a generic criteria was found. For example, for the subset \textit{Judicial\_Ethics}, one  criteria was described as follows by different groups: transparency, professional neutrality, conflict of interest. The last one  was chosen to encompass all the others. The final criteria list is summarized in Table~\ref{tab:criteria_groups}.

\paragraph{Agreements on final decisions.}
After verifying the consistency ratio, we further evaluated the inter-group agreement for each subset based on the final annotation decisions.
For each QA item, we measured the Agreement Ratio (AR), defined as the proportion of groups selecting the same (majority-voted) label.
Most QA items showed high agreement (AR >= 0.6), indicating strong consensus across groups, while only a few exhibited moderate variation (AR < 0.5).
This provides a concrete measure of consensus across annotator groups, complementing the internal consistency check.

We additionally compared the inter-group agreement ratios with those derived from the LegalBench gold standard.
As shown in Table~\ref{tab:gold_vs_human}, the average agreement reached 80\%, demonstrating strong alignment between group-level judgments and the gold-standard annotations. The Decision section, however, exhibited a lower agreement ratio, likely because its multiple-choice format made it more difficult for annotators to reach consensus.

\begin{table}[h]
\centering
\small
\begin{tabular}{l c}
\hline
\textbf{Subset} & \textbf{Agreement Ratio (\%)} \\
\hline
Abercrombie      & 86.7 \\
Judicial\_Ethics & 93.3 \\
Common\_Law        & 93.3 \\
Decision\_Section         & 60.0 \\
\hline
\end{tabular}
\caption{Agreement ratios between gold \textsc{LegalBench} and gold \textsc{Legal-AHP}. 
}
\label{tab:gold_vs_human}
\end{table}

\section{Methodology}  \label{sec:exp}
\subsection{AHP-based  Strategies}

Let a decision task be defined by a natural language question and a set of candidate alternatives $\mathcal{A} = \{a_1, a_2, \dots, a_n\}$. The goal is to identify the best alternative $a^*$ using a set of the criteria $\mathcal{C} = \{c_1, c_2, \dots, c_m\}$ and associated importance weights $\mathbf{w} = [w_1, \dots, w_m]$:
\[
a^* = \arg\max_{a_i \in \mathcal{A}} \sum_{j=1}^m w_j \cdot s(a_i, c_j)
\]
where $s(a_i, c_j)$ is the score of alternative $a_i$ with respect to criterion $c_j$.  We designed three methods to arrive at the decision $a^*$. Their corresponding prompts are illustrated in Appendix \ref{sec:prompt}.


\paragraph{(1) AHP Instruction.}
Unlike prior work that employed only partial elements of AHP such as hierarchical structuring of prompts~\citep{xu2024toward} or localized pairwise comparisons within specific decision stages ~\citep{svoboda2024enhancing, canchatgpt2023} our instruction set requires the LLM to execute the \textit{entire AHP decision-making process}.  
The model is guided through defining the problem, identifying key criteria, estimating relative importance via structured pairwise comparisons, and synthesizing results into a transparent and interpretable ranking of alternatives, as shown in Table~\ref{ahpInstruction}.

\paragraph{(2) Single-Agent AHP Framework.}

We further refine the AHP instruction by implementing it as a structured three-step pipeline, executed through multiple turns of interaction, as shown in Table~\ref{singleagent}. Specifically,  
(i) the model first generates a set of key decision criteria from the given question and alternatives;  
(ii) it then performs pairwise comparisons to derive relative weights for these criteria; and  
(iii) finally, it computes and ranks the alternatives to identify the optimal choice.  

Each stage is conditioned on the outputs of the previous one, maintaining a continuous dialogue history that captures the evolving reasoning process. During weighting and aggregation, the assistant explicitly refers back to its earlier responses, ensuring internal consistency and interpretability of the final recommendation.  
This multi-turn prompting design 
enhances the transparency andexplainability of the model’s decision-making compared to the single-turn AHP instruction.

\paragraph{(3) Multi-Agent AHP Framework.}
We propose a multi-agent AHP framework that enables structured, interpretable, and context-aware decision-making with LLMs.
As illustrated in Figure~\ref{fig:my_wide_figure} (Appendix~\ref{appendix:ahp}) and the corresponding prompts in Tables ~\ref{agentPrompts} and \ref{autoagentPrompts} (Appendix~\ref{sec:prompt}),
the framework consists of two key components, which collaboratively decompose the AHP workflow and simulate expert-level reasoning.

(a) \textit{Information Block.} It summarizes contextual relations between the decision problem and alternatives. The Retrieval/Summary agent generates a structured summary $I$ that contextualizes how each alternative connects to the decision intent:
\[
I = \texttt{SummarizeContext}(\text{Q}, \mathcal{A})
\]
Rather than retrieving external sources, this agent performs in-situ inference using only the provided inputs, enabling robust performance even in constrained or sensitive domains.
\\

(b) \textit{AHP Block.} Through pairwise comparisons, it derives priority weights for criteria and alternatives, and aggregates them into a final ranking (see Appendix~\ref{appendix:ahp} for details).   
It is implemented as a multi-agent module composed of three specialized agents: 

-- \textit{Agent 1: Criteria Generation.}
Given the decision question, alternatives, and the summary $I$ produced by the Information Block, this agent extracts a set of the criteria $\mathcal{C}$ that reflect the dimensions relevant for comparing the alternatives:
\[
\mathcal{C} = \texttt{GenerateCriteria}(\text{Q}, \mathcal{A}, I)
\]
Each criterion $c_j$ is phrased in natural language (e.g., “distinctiveness”, “functionality”) and designed to support transparent downstream evaluation.

-- \textit{Agent 2: Pairwise Weights.}
To assess the relative importance of each criterion, this agent constructs a pairwise comparison matrix $M \in \mathbb{R}^{m \times m}$:
\[
M_{ij} = \texttt{LLMCompare}(c_i, c_j \mid \text{Q}, \mathcal{A}, I)
\]
\[
w_j = \frac{1}{m} \sum_{i=1}^{m} \frac{M_{ij}}{\sum_{k=1}^{m} M_{ik}}
\]

-- \textit{Agent 3: Decision Making.}
The agent computes the final score of each alternative using the weighted criteria:
\[
\texttt{Score}(a_i) = \sum_{j=1}^m w_j \cdot s(a_i, c_j)
\]
It selects the top-ranked alternative and produces a natural language explanation reflecting how each criterion contributed to the decision:
\[
a^* = \arg\max_{a_i} \texttt{Score}(a_i)
\]
\[
\texttt{Explain}(a^*) = \texttt{LLMExplain}(a^*, C, w,Q)
\]


In this third strategy,  
agents can either be:

\begin{itemize}
    \item Manually constructed (\textbf{Multi-Agent (Manual)}) where the roles, responsibilities, and prompt templates of each agent are explicitly defined based on AHP to execute key stages of the AHP workflow, including information organization, criterion generation, weight estimation, and decision aggregation.

    \item Automatically generated (\textbf{Multi-Agent (Auto)}). Here the agent architecture is no longer statically specified. Instead, a centrally coordinated meta-agent dynamically decomposes the decision task and automatically instantiates a set of specialized agents according to the structure and complexity of the problem. Each generated agent is assigned a task-specific role, enabling collaborative execution of the AHP reasoning process. We used the \textit{CaptainAgent} framework~\citep{song2025adaptiveinconversationteambuilding} to implement this second setting.

\end{itemize}

Unlike the manually constructed agents, generated agents are created  in a model-driven manner, allowing adaptive organization and coordination during inference. 
Despite this difference in agent generation, both settings share the same decision objective, criterion weighting scheme, and aggregation formulation, ensuring a fair and controlled comparison between the two settings.





\subsection{Experimental Settings}

We evaluate our framework on four representative LLMs, including three open-source models, \textbf{Llama-3-8B-Instruct}~\cite{grattafiori2024llama3}, \textbf{Qwen2.5-7B-Instruct}~\cite{qwen25}, and \textbf{Mistral-7B-Instruct-v0.2}~\cite{jiang2023mistral7b},as well as the proprietary \textbf{GPT-4o-mini}~\cite{gpt4tr}.%
\footnote{We also evaluated \textbf{AdaptLLM/law-chat}~\cite{cheng2024adapting}, but omit it due to consistently lower performance.}

Each model is prompted with one of our AHP-based strategies and compared against a \textbf{Non-AHP} baseline using standard few-shot prompting from LegalBench. All strategies are provided with identical decision inputs and the same set of 5--8 few-shot examples to ensure fair comparison.

For the automatically generated multi-agent setting, we focus on the strongest open-source model (Qwen2.5-7B-Instruct) and the proprietary GPT-4o-mini to further investigate the effectiveness of AHP in a collaborative setting.

All models are evaluated with identical prompts and generation parameters. We fix the temperature to 0.7 and report accuracy averaged over three runs. Additional results, including macro-F1 scores, are reported in Appendix~\ref{tab:model_comparison_f1_accuracy}.

\section{Results}\label{sec:res}

\begin{table*}[h!]
\centering
\small
\begin{tabular}{@{}llllll@{}}
\toprule
\textbf{Dataset} & \textbf{Baseline} & \textbf{Mistral} & \textbf{LLAMA3} & \textbf{Qwen2.5-7B} & \textbf{GPT-4o-mini} \\

\midrule
Overall         & Non-AHP &  \textbf{68.3\%}&  \textbf{68.3\%} &  \textbf{71.7\%}  &  73.3\% \\
 & AHP-Instruction  & 66.7\% & 53.3\%  &63.3\% & 78.3\%\\
 & Single-Agent  & 43.3\%  &43.3\% & 66.7\%& 70.0\%\\
 & Multi-Agent (Manual) & 36.7\%  &66.7\% & 68.3\%& 61.7\%\\
  & Multi-Agent (Auto) &  - & - &65.0\%  & \textbf{81.7}\%\\
\bottomrule
\end{tabular}
\caption{Overall  model accuracy when evaluated on \textsc{LEGAL-AHP} (RQ1). The best score  is shown in \textbf{bold}. Detailed scores per  subset are presented in Table~\ref{tab:model_comparison_f1_accuracy}.}
\label{tab:model_comparison}
\end{table*}

We conduct experiments to answer four research questions in the following  subsections respectively: (RQ1) \textit{How does  Non-AHP and AHP-based strategies perform when compared to the  original \textsc{LegalBench} where legal problems  are framed as a QA task?} (RQ2) \textit{How effective and faithful are the intermediate decision criteria generated by our AHP prompting when compared to \textsc{Legal-AHP} gold criteria?} (RQ3) \textit{How does the AHP strategies output compare with human final decisions as given by \textsc{Legal-AHP}?} (RQ4) \textit{Do AHP-based strategies generalize to other decision problems, demonstrating portability beyond the legal domain?}




\subsection{LLMs vs. \textsc{LegalBench} Decisions} 

\paragraph{Non-AHP vs. AHP-based Methods.}
As shown in Table~\ref{tab:model_comparison}, the Non-AHP baseline provides strong and stable performance across models, with overall accuracies ranging from 68.3\% to 73.3\%. Among AHP-based methods, performance varies substantially by model. Notably, \textbf{GPT-4o-mini} consistently benefits from AHP prompting: AHP-Instruction improves accuracy from 73.3\% to 78.3\%, and the automatically generated Multi-Agent setting further raises performance to 81.7\%, the best result observed. In contrast, \textbf{Mistral} and \textbf{LLAMA3} experience notable performance degradation under AHP-Instruction and Single-Agent prompting, indicating limited robustness to structured decision prompts. \textbf{Qwen2.5} remains comparatively stable, maintaining near-baseline accuracy under Single-Agent and Multi-Agent settings.

\paragraph{Comparison within AHP-based Methods.}
Table~\ref{tab:model_comparison_f1_accuracy} illustrates the relative behavior of different AHP-based variants across models. AHP-Instruction shows the highest potential gains but primarily benefits stronger models, most notably GPT-4o-mini. Single-Agent prompting exhibits mixed and model-dependent effects, often degrading weaker models while remaining viable for Qwen2.5 and GPT-4o-mini. Multi-Agent prompting further amplifies this divergence: the manually constructed variant yields limited or inconsistent improvements, whereas the automatically generated Multi-Agent configuration provides the most stable and consistent gains for GPT-4o-mini. Overall, these results indicate that the effectiveness of AHP-based methods depends critically on both model capacity and the level of agent coordination.

\subsection{Evaluation of Generated Decision Criteria}

\begin{table*}[htbp]
\centering
\small 
\begin{tabular}{@{}lllll@{}}
\toprule
\textbf{Methods} & \textbf{Mistral} & \textbf{LLAMA3} & \textbf{Qwen2.5} & \textbf{GPT-4o-mini}\\

\midrule
AHP-Instruction & 9 & 8 & 5 & 13 \\
\midrule
Single-Agent & \textbf{47} & \textbf{53} &32 &\textbf{40}\\
\midrule
Multi-Agent (Manual) & 19 & 14 & \textbf{38} & 22\\
\bottomrule
\end{tabular}
\caption{Number of instances (out of 75) where each method’s generated decision criteria were judged as most aligned with human reasoning (RQ2). }
\label{tab:llm_judge}
\end{table*}

 Motivated by the idea that structured processes such as AHP can guide models toward more meaningful and transparent dimensions, (RQ2)   examines whether decision criteria generated by different methods align with human reasoning. 

As references, we summarize the representative criteria identified by annotators for each subset (Table~\ref{tab:criteria_groups}). Using an LLM-as-a-judge strategy (Qwen3-30B-A3B), the LLM extracts criteria from each method’s output, compares them with the human set, and selects the closest match. Table~\ref{tab:llm_judge} shows how often each method is selected as the best-aligned option across different subsets and models. 

The Single-Agent stands out as the most frequently chosen approach (e.g., 47 for Mistral, 53 for LLaMA3, 40 for GPT-4o-mini), suggesting that its broad coverage makes it more likely to capture human-relevant dimensions. In contrast, the AHP Instruction  is selected less often (e.g., 13 for GPT-4o-mini). This indicates that in one-shot settings models find it difficult to follow multi-step AHP instructions consistently, leading to less stability in criteria generation, even though they tend to achieve higher accuracy than other methods. The AHP Multi-Agent shows competitive results, particularly with Qwen2.5 (38), highlighting that dividing roles among agents can improve alignment when the base model is capable of effective coordination.

\subsection{LLMs vs. \textsc{Legal-AHP} Decisions} 

\begin{table*}[h!]
\centering
\begin{tabular}{lcccc}
\hline
\textbf{Method} & \textbf{Mistral} & \textbf{LLAMA3} & \textbf{Qwen2.5} & \textbf{GPT-4o-mini} \\
\hline
AHP-Instruction    & \textbf{61.3\%} & \textbf{48.0\%} & \textbf{69.3\%} & \textbf{80.0\%} \\
Single-Agent       & 38.7\%          & 47.0\%          & 64.0\%          & 61.0\% \\
Multi-Agent (Auto) & --              & --              & 65.3\%          & 69.3\% \\
\hline
\end{tabular}
\caption{Accuracy of different models evaluated against human-annotated \textsc{Legal-AHP} decisions (RQ3).}
\label{tab:anotated_accuracy}
\end{table*}

As discussed in (RQ2), Single-Agent prompting produces decision criteria that are more closely aligned with human annotations, whereas AHP-Instruction can be less stable when following complex multi-step instructions.
Motivated by this observation, we further evaluate how these prompting strategies translate into final decision accuracy when compared against human-annotated \textsc{Legal-AHP} labels.
The results are summarized in Table~\ref{tab:anotated_accuracy}.

Compared with the automatic evaluation in RQ1, several consistent trends emerge.
GPT-4o-mini achieves the strongest overall performance under AHP-Instruction, reaching 80.0\% accuracy and improving over its automatic evaluation result (76.0\%), which indicates closer alignment with human judgments.
Qwen2.5 exhibits relatively stable performance across prompting strategies.
In contrast, Mistral and LLAMA3 show limited performance gains under all evaluated settings.
Based on these observations, we do not further evaluate the Multi-Agent (Auto) framework on Mistral and LLAMA3, and instead focus on Qwen2.5 and GPT-4o-mini as representative open-source and proprietary models, respectively, for the multi-agent experiments.

Overall, the comparison between (RQ1) and (RQ3) suggests that although Single-Agent prompting aligns better with human reasoning at the criteria level (RQ2), AHP-based strategies yield more stable improvements in final decision accuracy, particularly for stronger models.

\subsection{Portability to Other Decision Problems}

To examine whether AHP-based strategies generalize beyond the legal domain, we evaluate our framework on two real-world ranking tasks: the \textit{QS World University Rankings} and the \textit{U.S. News Best Global Universities Rankings}.
For each benchmark, we restrict the candidate set to the top-10 universities from the official rankings and ask the model to re-rank them under identical conditions.

Given this fixed candidate pool, we compare a zero-shot baseline (Non-AHP), which directly outputs a ranking, with an AHP-based strategy.
Motivated by its strong performance in the legal domain, we adopt the automatically generated multi-agent AHP framework (Multi-Agent Auto), where the model explicitly generates decision criteria, assigns relative importance via pairwise comparisons, and aggregates weighted scores to produce the final ranking.

Tables~\ref{tab:qs_ranking} and~\ref{tab:us_ranking} in Appendix \ref{sec:ranking} report Kendall’s Tau between model-generated rankings and the official reference rankings.
On the QS benchmark, Multi-Agent (Auto) improves ranking correlation for GPT-4o-mini, while Non-AHP remains slightly stronger for Qwen2.5-7B-Instruct.
On the U.S. News benchmark, Multi-Agent (Auto) yields a substantial improvement for Qwen2.5-7B-Instruct, whereas GPT-4o-mini performs better under the Non-AHP baseline.
We further analyze whether the generated criteria align with the ones used by the official ranking. Tables~\ref{tab:QS_rank_by_position} and~\ref{tab:usnews_rank_by_position}   show that AHP-based rankings produce more stable top-tier ordering, particularly among highly ranked universities, whereas zero-shot rankings exhibit larger fluctuations.
These results suggest that structured AHP reasoning improves robustness and generalization in real-world ranking tasks.
Overall, these results indicate that AHP-based reasoning exhibits meaningful portability to non-legal ranking tasks, enabling LLMs to perform structured multi-criteria decision-making beyond the legal domain.

\section{Discussion} \label{sec:discu}

The integration of structured decision-making frameworks into LLM prompting substantially improves interpretability and alignment with expert judgments.
However, these benefits are not uniform across models.
Stronger LLMs, such as GPT-4o-mini and Qwen2.5, are better able to internalize and exploit explicit reasoning scaffolds, whereas weaker models struggle to reliably follow multi-step analytical procedures.
These observations suggest that the effectiveness of structured prompting is closely tied to the underlying reasoning capacity and instruction-following stability of the model.

An important finding is that LLMs can partially automate the Analytic Hierarchy Process.
Across experimental settings, models are capable of generating meaningful intermediate criteria and producing interpretable final decisions, but their reliability decreases as the depth and coordination requirements of reasoning increase.
Different AHP realizations exhibit complementary strengths: single-agent formulations align more closely with human reasoning at the criterion level, while AHP-guided aggregation yields more stable improvements in final decision accuracy.
Together, these results indicate that effective automation depends not only on prompt design, but also on how reasoning steps are decomposed, ordered, and coordinated within the decision pipeline.

Beyond the legal domain, the proposed framework demonstrates meaningful generalization.
As shown in (RQ4), the same AHP-based reasoning pipeline transfers effectively to non-legal decision problems, such as real-world ranking tasks, without relying on domain-specific knowledge.
Such portability is enabled by the modular structure of the framework, which decouples criterion generation, weighting, and aggregation, allowing the same reasoning structure to be reused across domains.

Finally, systematic differences emerge across task formats.
Binary decision problems are generally easier for LLMs, whereas multiple-choice and comparative settings benefit most from explicit structure and multi-criteria aggregation.
Taken together, these findings suggest that while structured decision frameworks can significantly enhance transparency and reliability, achieving fully robust and interpretable reasoning automation in complex, high-stakes, and cross-domain settings remains an open challenge.

\section{Conclusion}

This paper proposes an agent-based framework that integrates the Analytic Hierarchy Process (AHP) with large language models to support transparent and interpretable multi-criteria decision-making.
By structuring reasoning into criterion generation, weighting, and aggregation, the framework transforms opaque model outputs into auditable decision processes. Experiments on the \textsc{Legal-AHP} benchmark demonstrate that AHP-based prompting enables LLMs to better reflect expert decision logic at the level of intermediate reasoning. This is particularity salient for the multi-agent strategy where 
the resulting decisions are substantially more interpretable and more closely aligned with human analytical reasoning. 
We further show that the framework generalizes beyond the legal domain to real-world ranking tasks without relying on domain-specific knowledge.

Future work will explore improved agent coordination, symbolic consistency verification, and extensions to broader high-stakes domains where transparency and accountability are critical.

\section*{Ethical Considerations}
The data used to build \textsc{Legal-AHP} are composed of question–answer pairs taken from publicly available datasets accessible to the research community, which do not contain any abusive content or privacy issues.

Regarding the annotation campaign, the annotators were Master’s students in law, and their participation was part of their academic training. The dataset will be anonymized prior to its release to ensure privacy protection.

It is important to note that decision-making in the legal domain is highly context dependent. The same law or decree may lead to different judgments depending on the specific circumstances of each case. In such situations, even a decision tree cannot capture the underlying reasoning, making full automation of legal reasoning impractical. Therefore, our aim is not to develop large language models that replace legal experts in complex settings, but rather to assist them as tools that help reduce cognitive load, improve analytical efficiency, and support fair and transparent decision-making.


\section*{Limitations}

This study has several limitations.First, the effectiveness of AHP-based strategies depends on the reasoning capability of the underlying language models and the stability of multi-agent coordination. As observed in our experiments, models with stronger instruction-following and reasoning abilities benefit more consistently from structured AHP prompting, whereas weaker models may struggle with multi-step or multi-agent reasoning. Future work should investigate adaptive coordination and communication mechanisms to improve robustness across different model capacities.

Second, the annotation process relies on expert-defined criteria and pairwise comparisons, which inevitably introduce a degree of subjectivity. Although consistency checks were applied to filter unreliable annotations, differences in annotators’ expertise and interpretation can still affect the resulting criteria and weights. Exploring hybrid human–model annotation pipelines or partially automated criterion generation may help mitigate this limitation while preserving annotation quality.

Finally, the \textsc{Legal-AHP} dataset is relatively small due to the complexity of AHP-based annotation. Constructing the dataset required multiple expert annotators to define consistent criteria, perform pairwise comparisons, and provide detailed reasoning justifications. This process is inherently time-consuming and cognitively demanding, limiting scalability but also making the dataset a high-quality benchmark for studying structured decision-making and interpretability.



\bibliography{custom}

\begin{thebibliography}{34}
\providecommand{\natexlab}[1]{#1}

\bibitem[{Badri(2001)}]{badri2001ahp}
Masood~A Badri. 2001.
\newblock A combined ahp-gp model for quality control systems.
\newblock \emph{Omega}, 29(5):387--406.

\bibitem[{Brans and De~Smet(2005)}]{brans2005promethee}
Jean-Pierre Brans and Yves De~Smet. 2005.
\newblock Promethee methods.
\newblock In \emph{Multiple criteria decision analysis: state of the art
  surveys}, pages 187--219. Springer.

\bibitem[{Cheng et~al.(2024)Cheng, Huang, and Wei}]{cheng2024adapting}
Daixuan Cheng, Shaohan Huang, and Furu Wei. 2024.
\newblock \href {https://openreview.net/forum?id=y886UXPEZ0} {Adapting large
  language models via reading comprehension}.
\newblock In \emph{The Twelfth International Conference on Learning
  Representations}.

\bibitem[{Cloud(2024)}]{qwen25}
Alibaba Cloud. 2024.
\newblock \href {https://arxiv.org/abs/2412.15115} {Qwen2.5 technical report}.
\newblock ArXiv preprint arXiv:2412.15115.

\bibitem[{Doe and Liu(2023)}]{canchatgpt2023}
Jane Doe and Han Liu. 2023.
\newblock Can chatgpt serve as a multi-criteria decision maker?
\newblock \emph{Proceedings of the 2023 Conference on Decision Science}.

\bibitem[{Eigner and H{\"a}ndler(2024)}]{eigner2024determinants}
Eva Eigner and Thorsten H{\"a}ndler. 2024.
\newblock \href {https://arxiv.org/abs/2402.17385} {Determinants of
  llm-assisted decision-making}.
\newblock \emph{arXiv preprint arXiv:2402.17385}.

\bibitem[{Forman and Gass(2001)}]{forman2001ahp}
Ernest~H Forman and Saul~I Gass. 2001.
\newblock The analytic hierarchy process—an exposition.
\newblock \emph{Operations Research}, 49(4):469--486.

\bibitem[{Govindan and Jepsen(2016)}]{govindan2016electre}
Kannan Govindan and Martin~Brandt Jepsen. 2016.
\newblock Electre: A comprehensive literature review on methodologies and
  applications.
\newblock \emph{European Journal of Operational Research}, 250(1):1--29.

\bibitem[{Grattafiori et~al.(2024)Grattafiori, Dubey, and
  et~al.}]{grattafiori2024llama3}
Aaron Grattafiori, Abhimanyu Dubey, and et~al. 2024.
\newblock \href {https://arxiv.org/abs/2407.21783} {The llama 3 herd of
  models}.
\newblock \emph{arXiv preprint arXiv:2407.21783}.

\bibitem[{Ho et~al.(2014)Ho, Rabah, Nowakowski, and
  Estraillier}]{tracebased2021}
Hoang~Nam Ho, Mourad Rabah, Samuel Nowakowski, and Pascal Estraillier. 2014.
\newblock A process for trace-based criteria weighting in multiple criteria
  decision making.
\newblock \emph{Journal of Software}.

\bibitem[{Ho(2008)}]{ho2008integrated}
William Ho. 2008.
\newblock An integrated analytic hierarchy process and its applications to
  service selection.
\newblock \emph{European Journal of Operational Research}, 186(1):211--228.

\bibitem[{Holzenberger et~al.(2023)Holzenberger, Zheng, Guha
  et~al.}]{holzenberger2023legalbench}
Nils Holzenberger, Cheng Zheng, Tanmay Guha, and 1 others. 2023.
\newblock Legalbench: A collaboratively built benchmark for measuring legal
  reasoning in large language models.
\newblock \emph{arXiv preprint arXiv:2308.11462}.

\bibitem[{Jiang and et~al.(2023)}]{jiang2023mistral7b}
Albert~Q. Jiang and et~al. 2023.
\newblock \href {https://arxiv.org/abs/2310.06825} {Mistral 7b}.
\newblock \emph{arXiv preprint arXiv:2310.06825}.

\bibitem[{Khosla and Singh(2021)}]{khosla2021fuzzy}
Ankit Khosla and Ram~Kumar Singh. 2021.
\newblock A fuzzy ahp approach for prioritizing indicators of sustainable
  manufacturing in the indian context.
\newblock \emph{Journal of Cleaner Production}, 281:124715.

\bibitem[{Labib(2011)}]{labib2011decision}
Ashraf~W Labib. 2011.
\newblock A decision analysis model for maintenance policy selection using ahp.
\newblock \emph{International Journal of Production Economics},
  132(2):174--182.

\bibitem[{Lu et~al.(2024{\natexlab{a}})Lu, Li, Takeuchi, and
  Kashima}]{xu2024toward}
Xiaotian Lu, Jiyi Li, Koh Takeuchi, and Hisashi Kashima. 2024{\natexlab{a}}.
\newblock Ahp-powered llm reasoning for multi-criteria evaluation of open-ended
  responses.
\newblock \emph{Findings of EMNLP}.

\bibitem[{Lu et~al.(2024{\natexlab{b}})Lu, Hu, Foroosh, Jin, and
  Liu}]{lu2024strux}
Yiming Lu, Yebowen Hu, Hassan Foroosh, Wei Jin, and Fei Liu.
  2024{\natexlab{b}}.
\newblock \href {https://arxiv.org/abs/2410.12583} {Strux: An llm for
  decision-making with structured explanations}.
\newblock \emph{arXiv preprint arXiv:2410.12583}.

\bibitem[{Munier et~al.(2021)Munier, Hontoria et~al.}]{munier2021uses}
Nolberto Munier, Eloy Hontoria, and 1 others. 2021.
\newblock Uses and limitations of the ahp method.

\bibitem[{OpenAI et~al.(2023)}]{gpt4tr}
OpenAI and 1 others. 2023.
\newblock \href {https://arxiv.org/abs/2303.08774} {Gpt-4 technical report}.
\newblock \emph{arXiv preprint arXiv:2303.08774}.

\bibitem[{Opricovic and Tzeng(2007)}]{opricovic2007extended}
Serafim Opricovic and Gwo-Hshiung Tzeng. 2007.
\newblock Extended vikor method in comparison with outranking methods.
\newblock \emph{European journal of operational research}, 178(2):514--529.

\bibitem[{Papasotiriou et~al.(2024)Papasotiriou, Sood, Reynolds, and
  Balch}]{LLM-Invest}
Kassiani Papasotiriou, Srijan Sood, Shayleen Reynolds, and Tucker Balch. 2024.
\newblock \href {https://doi.org/10.1145/3677052.3698694} {Ai in investment
  analysis: Llms for equity stock ratings}.
\newblock In \emph{Proceedings of the 5th ACM International Conference on AI in
  Finance}, ICAIF '24, page 419–427, New York, NY, USA. Association for
  Computing Machinery.

\bibitem[{Papathanasiou and Ploskas(2018)}]{papathanasiou2018topsis}
Jason Papathanasiou and Nikolaos Ploskas. 2018.
\newblock Topsis.
\newblock In \emph{Multiple criteria decision aid: Methods, examples and python
  implementations}, pages 1--30. Springer.

\bibitem[{Park et~al.(2025)Park, Oh, Gao, and Kwon}]{park2025enhancing}
Haeun Park, Hyunjoo Oh, Feng Gao, and Ohbyung Kwon. 2025.
\newblock Enhancing analytic hierarchy process modelling under uncertainty with
  fine-tuning llm.
\newblock \emph{Expert Systems}, 42(6):e70051.

\bibitem[{Ribeiro et~al.(2016)Ribeiro, Singh, and Guestrin}]{ribeiro2016lime}
Marco~Tulio Ribeiro, Sameer Singh, and Carlos Guestrin. 2016.
\newblock \"why should i trust you?\": Explaining the predictions of any
  classifier.
\newblock In \emph{Proceedings of the 22nd ACM SIGKDD International Conference
  on Knowledge Discovery and Data Mining}, pages 1135--1144.

\bibitem[{Saaty(1980)}]{saaty1980ahp}
Thomas~L Saaty. 1980.
\newblock \emph{The Analytic Hierarchy Process}.
\newblock McGraw-Hill.

\bibitem[{Saaty(2005)}]{saaty2005ahp}
Thomas~L. Saaty. 2005.
\newblock \href {https://doi.org/10.1007/0-387-23081-5_9} {The analytic
  hierarchy and analytic network processes for the measurement of intangible
  criteria and for decision-making}.
\newblock In Jos{\'e} Figueira, Salvatore Greco, and Matthias Ehrgott, editors,
  \emph{Multiple Criteria Decision Analysis: State of the Art Surveys},
  International Series in Operations Research \& Management Science, pages
  345--405. Springer, New York, NY.

\bibitem[{Sha et~al.(2023)Sha, Zhang, Wang, Guo, Zhang, and
  He}]{sha2023languagempc}
Yicheng Sha, Yifan Zhang, Rui Wang, Yu~Guo, Yue Zhang, and Xiaodong He. 2023.
\newblock \href {https://arxiv.org/abs/2310.03026} {Languagempc: Large language
  models as decision makers for autonomous driving}.
\newblock \emph{arXiv preprint arXiv:2310.03026}.

\bibitem[{Shapley(1953)}]{shapley1953value}
Lloyd~S Shapley. 1953.
\newblock A value for n-person games.
\newblock \emph{Contributions to the Theory of Games}, 2:307--317.

\bibitem[{Simon(1960)}]{simon1960new}
Herbert~A Simon. 1960.
\newblock The new science of management decision.

\bibitem[{Soltanpanah et~al.(2018)Soltanpanah, Yazdani, and
  Kabir}]{soltanpanah2018application}
Hossein Soltanpanah, Morteza Yazdani, and Ghulam Kabir. 2018.
\newblock Application of ahp and fuzzy ahp for project selection under
  uncertainty: A case study.
\newblock \emph{Technological and Economic Development of Economy},
  24(2):674--691.

\bibitem[{Song et~al.(2025)Song, Liu, Zhang, Zhang, Luo, Wang, Wu, and
  Wang}]{song2025adaptiveinconversationteambuilding}
Linxin Song, Jiale Liu, Jieyu Zhang, Shaokun Zhang, Ao~Luo, Shijian Wang,
  Qingyun Wu, and Chi Wang. 2025.
\newblock \href {https://arxiv.org/abs/2405.19425} {Adaptive in-conversation
  team building for language model agents}.
\newblock \emph{Preprint}, arXiv:2405.19425.

\bibitem[{Stelmach(2025)}]{stelmach2025evaluating}
Alexander Stelmach. 2025.
\newblock Evaluating ai judgements: A case study of llms in product
  development.

\bibitem[{Sun et~al.(2025)Sun, Huang, and Pompili}]{sun2025llm_madm}
Chuanneng Sun, Songjun Huang, and Dario Pompili. 2025.
\newblock \href {https://doi.org/10.1109/LRA.2025.3535588} {Llm-based
  multi-agent decision-making: Challenges and future directions}.
\newblock \emph{IEEE Robotics and Automation Letters}.

\bibitem[{Svoboda and Lande(2024)}]{svoboda2024enhancing}
Igor Svoboda and Dmytro Lande. 2024.
\newblock Enhancing multi-criteria decision analysis with ai: Integrating
  analytic hierarchy process and gpt-4.
\newblock \emph{arXiv preprint arXiv:2402.07404}.

\end{thebibliography}

\appendix

\section{Analytic Hierarchy Process (AHP)}
\begin{figure*}[htb!] 
    \centering
    \includegraphics[width=\textwidth] {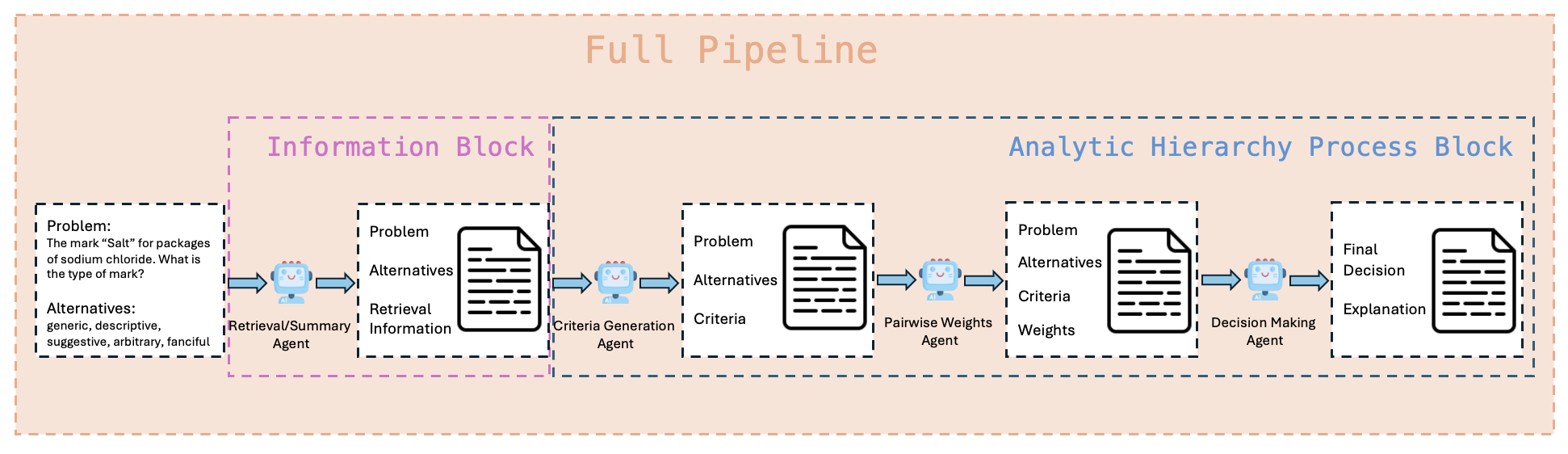} 
    \caption{Pipeline of our multi-agent decision-making framework 
    illustrated here on a  legal classification task from the   \textsc{Legal-AHP} subset aiming to determine the trademark category of the term \textit{“Salt”}.  
    }
    \label{fig:my_wide_figure}
\end{figure*}

\label{appendix:ahp}

According to its founder \citep{saaty1980ahp}, the Analytic Hierarchy Process (AHP) is built on three key principles: (i) hierarchical structuring of the decision problem, (ii) priority derivation through pairwise comparisons, and (iii) logical consistency checks. In decision-support contexts such as student accommodation allocation \citep{saaty2005ahp}, AHP is applied to derive the weights of the criteria before applying other multi-criteria ranking methods.

\begin{table*}[hbt!]
\centering

\begin{tabular}{ccl}
\hline
Intensity & Definition        & Explanation \\
\hline
1 & Equal importance   & Neither alternative is preferred \\
3 & Weak importance    & One alternative is slightly preferred \\
5 & Strong importance  & Clear preference of one over the other \\
7 & Very strong        & Strongly favored alternative \\
9 & Extreme            & Extremely more important \\
2,4,6,8 & Intermediate & Compromise values between the above \\
\hline
\end{tabular}
\caption{Saaty’s fundamental scale \citep{saaty1980ahp}.}
\label{tab:saaty_scale}
\end{table*}

\subsection*{Hierarchical structure}
The problem is decomposed into three levels: (i) the overall goal at the top, (ii) the criteria in the middle, and (iii) the alternatives at the bottom.

\subsection*{Pairwise comparisons}
The relative importance of each criterion is elicited using Saaty’s fundamental scale (Table~\ref{tab:saaty_scale}). From these judgments, a comparison matrix $A = [a_{ij}]$ is constructed, where
\[
    a_{ij} = \frac{w_i}{w_j}, \quad i,j = 1,\dots,n,
\]
and $w_i$ denotes the weight of criterion $i$.

\subsection*{Normalization and weight derivation}
The comparison matrix is normalized to obtain a matrix $B = [b_{ij}]$:
\[
    b_{ij} = \frac{a_{ij}}{\sum_{j=1}^n a_{ij}}.
\]
The weight of each criterion $i$ is then computed as the row average:
\[
    W_i = \sum_{j=1}^n b_{ij}.
\]

\subsection*{Consistency check}
To ensure judgments are logically consistent, the following quantities are calculated:
\begin{align}
    c_i &= \sum_{j=1}^n a_{ij} w_j, \\
    \lambda_{\max} &= \frac{1}{n} \sum_{i=1}^n \frac{c_i}{w_i}, \\
    CI &= \frac{\lambda_{\max} - n}{n - 1}, \\
    CR &= \frac{CI}{RI},
\end{align}
where $n$ is the number of criteria and $RI$ is the random index (Table~\ref{tab:random_index}). A pairwise comparison matrix is considered consistent if $CR \leq 0.1$.

\begin{table*}[hbt!]
\centering
\begin{tabular}{ccccccccccc}
\hline
$n$  & 1 & 2 & 3 & 4 & 5 & 6 & 7 & 8 & 9 & 10 \\
\hline
$RI$ & 0 & 0 & 0.58 & 0.90 & 1.12 & 1.24 & 1.32 & 1.41 & 1.45 & 1.49 \\
\hline
\end{tabular}
\caption{Random Index Values (RI) for different $n$ \citep{saaty1980ahp}.}
\label{tab:random_index}
\end{table*}

\subsection*{Manual Multi-Agent Construction}

Figure~\ref{fig:my_wide_figure} provides an overview of how the Analytic Hierarchy Process (AHP) is instantiated within our manually constructed multi-agent framework. The figure illustrates the end-to-end decision pipeline on a representative legal classification task from the \textsc{Legal-AHP} dataset, where the objective is to determine the trademark category of the term \textit{``Salt''}.

As depicted, the framework is organized into two core components. The Information Block first processes the decision question and candidate alternatives to extract and summarize task-relevant contextual information. This structured context is then forwarded to the AHP Block, which decomposes the decision-making process into criterion generation, pairwise weighting, and final aggregation. Each stage is handled by a dedicated agent with a predefined role, enabling a clear separation of responsibilities across the workflow.
 gb
Overall, the figure offers a high-level visualization of how information analysis and AHP-style structured reasoning are integrated in the manual multi-agent setting. It complements the detailed prompt templates and algorithmic formulations introduced in subsequent sections, and highlights the transparency and modularity of the proposed design.

\section{Details of the Annotation Procedure}

\subsection{Selected Subsets}
\label{appendix:subsets}

We select five subsets from \textsc{LegalBench}, each representing a distinct legal domain and decision-making format, as shown in Table~\ref{tab:legalbench_subsets}:

\begin{itemize}
    \item The \texttt{Abercrombie} task assesses trademark distinctiveness by classifying a mark as Generic, Descriptive, Suggestive, Arbitrary, or Fanciful.

    \item The \texttt{Judicial\_Ethics} task evaluates whether certain actions by New York State judges are ethically permissible, based on official advisory opinions.

    \item The \texttt{Common\_Law} task determines whether a given contract is governed by the Uniform Commercial Code (UCC) or by common law, depending on the contract's subject matter.

    \item The \texttt{Decision\_Section} task classifies paragraphs from judicial decisions into one of seven functional roles: Facts, Procedural History, Issue, Rule, Analysis, Conclusion, or Decree.

    \item  Finally, the \texttt{Privacy\_Policy} task asks whether a given privacy policy excerpt is relevant to answering a specific question, framed as a binary classification problem.

\end{itemize}

\begin{table*}[th]
\centering
\small
\begin{tabularx}{\textwidth}{|l|X|l|}
\hline
\textbf{Subtask} & \textbf{Example Question} & \textbf{Answer} \\ \hline
\texttt{Abercrombie} & The mark “Salt” for packages of sodium chloride. What is the type of mark? & Generic \\ \hline
\texttt{Judicial\_Ethics} &  Is a judge required to disclose a former law clerk's employment relationship when the former law clerk appears in an adversarial role within one year after his/her employment with the judge ended? & Yes \\ \hline
\texttt{Common\_Law} & Contract: Grace and Hudson form an agreement for the sale of a microwave. Is this contract governed by the UCC or the common law? & UCC \\ \hline
\texttt{Decision\_Section} & On appeal, Khochinsky challenges the district court's dismissal under the FSIA as well as the court's vacatur of the default. & Issue \\ \hline
\texttt{Privacy\_Policy} & Clause: Personal information such as purchase history shall not be further processed or used for any commercial purposes.
 \newline Question: is there a way to opt out of data sharing & Relevant \\ \hline
\end{tabularx}
\caption{Example question--answer pairs for the selected \textsc{LegalBench} subsets.}
\label{tab:legalbench_subsets}
\end{table*}

\subsection{Training Phase} \label{training}
We detail below how QA pairs in each subset were selected for the training phase:

\begin{itemize}
    \item \texttt{Abercrombie}: Training on the 4 first questions of each series. Total number of training questions$=$32.
    \item \texttt{Common\_Law}: Training on the 5 first questions of each series. Total number of training questions$=$75.
    \item \texttt{Decision\_Section}: Training on the 7 firsAccording to its foundet questions of each series. Total number of training questions$=$105.
    \item \texttt{Judicial\_Ethics}: Training on the 8 first questions of each series. Total number of training questions$=$120.
    \item \texttt{Privacy\_Policy}: Training on the 8 first questions of each series. Total number of training questions$=$120.
\end{itemize}

\subsection{AHP-Based Annotation Guidelines}
\label{appendix:ahp_details}
This appendix describes the procedure that annotators followed when applying the Analytic Hierarchy Process (AHP) to each question. In this setting, the goal is the question itself, the alternatives are the answer options, and the best alternative is the correct answer. The annotators carried out the following steps:

\begin{enumerate}
    \item \textbf{Identify the criteria:}  
    Define a set of the criteria that are relevant for determining the best answer. These criteria should capture different aspects of what makes an answer correct or appropriate (e.g., legal relevance, clarity, completeness).

    \item \textbf{Create pairwise comparison matrices:}  
    Construct pairwise comparison matrices to evaluate (a) the relative importance of the criteria and (b) the relative performance of the alternatives under each criterion. Use the standard AHP 1–9 scale, where 1 indicates equal importance and 9 indicates extreme importance. Use reciprocal values (e.g., 1/3, 1/5) when one item is less important than another.

    \item \textbf{Derive weights:}  
    Normalize each pairwise comparison matrix and average the rows to compute the priority weights. This produces (i) the weights for each criterion and (ii) the weights for each alternative under each criterion.

    \item \textbf{Evaluate alternatives:}  
    Combine the weights of criteria and alternatives to calculate an overall score for each alternative. The alternative with the highest overall score is considered the best answer to the question.
\end{enumerate}

These steps were provided to all student annotators to ensure consistency and rigor in the annotation process.

\subsection{Illustrative Example}

Table~\ref{tab:annotation_example} illustrates an example of the AHP-based annotation process, as applied by a group of annotators to the subset \textit{Decision\_Section}.

\begin{table*}[h]
\small
\centering
\begin{tabular}{|l|c|c|c|}
\hline
\textbf{Criteria} & \textbf{Consistency of lexical scope}  & \textbf{Level of precision } & \textbf{Subject} \\ \hline
Consistency of lexical scope & 1.00 & 0.20 & 0.14 \\ \hline
Level of precision & 5.00 & 1.00 & 0.20 \\ \hline
Subject & 7.00 & 5.00 & 1.00 \\ \hline
\end{tabular}
\caption{Criteria comparison matrix as given by annotators from one group for the subset \texttt{Decision\_Section}.}
\label{tab:annotation_example}
\end{table*}

\section{Detailed Results in Terms of F1-scores}

\begin{table*}[h!]
\centering
\small
\begin{tabular}{@{}llllll@{}}
\toprule
\textbf{Dataset} & \textbf{Baseline} & \textbf{Mistral} & \textbf{LLAMA3} & \textbf{Qwen2.5-7B} & \textbf{GPT-4o-mini} \\
\midrule
\multirow{4}{*}{Abercrombie}  & Non-AHP & \textbf{53.3\%}  &  \textbf{53.3\%} & \textbf{53.3\%} & \textbf{73.3\%} \\
 & AHP-Instruction  & 46.7\% & 40.0\% & 33.3\% & 60.0\%\\
 & Single-Agent & 40.0\%  & 13.3\%   & 46.7\% & 60.0\% \\
 & Multi-Agent(Manual)  & 26.7\%  &40.0\% & 40.0\%& 40.0\%\\
   & Multi-Agent(Auto) & -  & -  &40.0\%  & 66.7\%\\
\midrule
Judicial\_Ethics & Non-AHP & \textbf{86.7\%} & \textbf{73.3\%}& 66.7\% & 73.3\% \\
  & AHP-Instruction  & 73.3\% & \textbf{73.3\%}  & 60.0\% & \textbf{86.7\%}\\
 & Single-Agent  & 40.0\%  & 60.0\%   & 60.0\% & 66.7\% \\
 & Multi-Agent(Manual)  & 20.0\%  &\textbf{73.3\%} & \textbf{73.3\%}& 73.3\%\\ 
   & Multi-Agent(Auto) &  - & - &60.0\%  & 80.0\%\\
\midrule
Common\_Law      & Non-AHP& 80.0\% & 80.0\% & \textbf{100.0\%} & \textbf{100.0\%}\\
 & AHP-Instruction  & \textbf{86.7\%} & 60.0\%  &86.7\% & 93.3\%\\
 & Single-Agent  & 53.3\%  &\textbf{86.7\%} & \textbf{100.0\%}& 86.7\%\\
 & Multi-Agent(Manual)  & 80.0\%  &\textbf{86.7\%} & 93.3\%& 93.3\%\\ 
  & Multi-Agent(Auto) & -  & -  & 93.3\%  & \textbf{100.0\%}\\
\midrule
Decision\_Section  & Non-AHP& 53.3\%  & \textbf{66.7\%} & 66.7\% & 46.7\%  \\
 & AHP-Instruction  & \textbf{60.0}\% & 40.0\%  &\textbf{73.3\%} & 73.3\%\\
 & Single-Agent & 40.0\%  &13.3\% & 60.0\% & 66.7\%\\
 & Multi-Agent(Manual)  & 20.0\%  &46.7\% & 66.7\%& 40.0\%\\ 
  & Multi-Agent(Auto) & -  & -  & 66.7\% & \textbf{80.0}\%\\
\midrule\midrule

\multirow{4}{*}{Abercrombie}  & Non-AHP & \textbf{51.8\%}  &  44.5\% & 45.3\% & \textbf{71.3\%} \\
 & AHP-Instruction  & 37.7\% & 33.2\% & 40.4\% & 64.8\%\\
 & Single-Agent & 39.8\%  & 13.0\%   & \textbf{46.6\%} & 58.7\% \\
 & Multi-Agent(Manual) & 26.1\%  &\textbf{48.4\%} & 29.3\% & 38.5\%\\
     & Multi-Agent(Auto) & -  & -  &40.0\%  & 66.7\%\\
\midrule
Judicial\_Ethics & Non-AHP & \textbf{86.6\%} & 70.0\% & 64.1\% & 60.0\% \\
  & AHP-Instruction  & 77.8\% & 60.0\%  & 58.3\% & \textbf{86.6\%}\\
 & Single-Agent  & 39.9\%  & 58.3\%   & 59.8\% & 76.9\% \\
 & Multi-Agent(Manual) & 20.0\%  &\textbf{73.2\%} & \textbf{79.7\%}& 73.2\%\\ 
     & Multi-Agent(Auto) &  - & - &60.0\%  & 80.0\%\\
\midrule
Common\_Law      & Non-AHP & 80.0\% & 80.0\% & \textbf{100.0\%} & \textbf{100.0\%}\\
 & AHP-Instruction  & \textbf{85.0\%} & 25.0\%  &86.6\% & 89.9\%\\
 & Single-Agent & 34.8\% & \textbf{85.0\%}  &\textbf{100.0\%} & 86.1\%\\
 & Multi-Agent(Manual) & 59.3\%  &82.9\% & 92.8\%& 92.1\%\\ 
     & Multi-Agent(Auto) & -   & - &93.3\%  & \textbf{100.0}\%\\
\midrule
Decision\_Section  & Non-AHP & \textbf{48.2\%}  & \textbf{57.0\%} & 70.8\% & 46.5\%  \\
 & AHP-Instruction  & 40.2\% & 30.1\%  &71.3\% & 73.9\%\\
 & Single-Agent & 29.5\% & 16.7\%  &45.7\% & \textbf{76.2\%}\\
 & Multi-Agent(Manual) & 22.9\%  &45.3\% & \textbf{73.3\%}& 25.9\%\\ 
    & Multi-Agent(Auto) & -  & - &56.2\%  & 55.0\%\\

\midrule
Overall         & Non-AHP &  \textbf{66.7\%}&  \textbf{62.9\%}&  \textbf{70.1\%}  &  69.5\% \\
 & AHP-Instruction  & 60.2\% & 37.0\%  &64.2\% & \textbf{78.8\%}\\
 & Single-Agent  & 36.0\%  &43.3\% & 63.0\%& 74.5\%\\
 & Multi-Agent(Manual) & 32.1\%  &62.5\% & 68.8\%& 57.4\%\\
   & Multi-Agent(Auto) & -  & - &52.1\%  & 72.7\%\\
\bottomrule
\end{tabular}

\caption{
Comparison of model performance across datasets.
The upper block reports accuracy, while the lower block reports F1-score.
The best score in each setting is shown in \textbf{bold}.
}
\label{tab:model_comparison_f1_accuracy}
\end{table*}

Table~\ref{tab:model_comparison_f1_accuracy} presents the detailed F1-scores across datasets and model configurations. Overall, the Non-AHP baseline remains the most stable across all models, with scores ranging from 62.9\% to 70.1\%. In contrast, AHP-Instruction and Single-Agent strategies show greater variance, often enhancing stronger models while degrading weaker ones. \textbf{GPT-4o-mini} achieves the best overall performance, reaching 78.8\% under AHP and 74.5\% with the Single-Agent setup, both exceeding its baseline of 69.5\%. \textbf{Qwen2.5} also performs consistently well, attaining perfect F1 (100.0\%) on \textit{Common Law} and high scores on other tasks. Conversely, \textbf{Mistral} and \textbf{LLAMA3} exhibit instability under AHP (37.7\% and 37.0\%, respectively) but moderate recovery in Multi-Agent settings, such as LLAMA3’s 73.2\% on \textit{Judicial Ethics}. Dataset-wise, GPT-4o-mini leads on \textit{Abercrombie} (71.3\%) and \textit{Judicial Ethics} (86.6\%), while both GPT-4o-mini and Qwen2.5 reach 100.0\% on \textit{Common Law}. On \textit{Decision Section}, GPT-4o-mini achieves its highest F1 (76.2\%) with the Single-Agent method, and Qwen2.5 peaks under Multi-Agent (73.3\%). Overall, Non-AHP offers the most reliable baseline, whereas AHP and Single-Agent strategies can significantly boost advanced models like GPT-4o-mini and Qwen2.5 but tend to destabilize smaller models such as Mistral and LLAMA3.

\section{Portability to Ranking Decision Problems}
\label{sec:ranking}

Tables~\ref{tab:qs_ranking} and~\ref{tab:us_ranking} report Kendall’s Tau scores on the QS World University Rankings and the U.S. News Best Global Universities Rankings.
We evaluate both a zero-shot Non-AHP baseline and the proposed Auto Multi-Agent AHP framework under an identical top-10 candidate setting.

Overall, AHP-based structured reasoning leads to more consistent rankings when the underlying model can reliably follow multi-step instructions.
On the QS benchmark, Auto Multi-Agent AHP improves performance for GPT-4o-mini, while showing comparable results to the Non-AHP baseline for Qwen2.5.
On the U.S. News benchmark, the AHP framework substantially boosts performance for Qwen2.5, indicating that explicit criteria decomposition and aggregation can effectively reduce ranking noise for weaker models.

Position-level comparisons in Tables~\ref{tab:QS_rank_by_position} and~\ref{tab:usnews_rank_by_position} further show that AHP-based rankings produce more stable top-tier ordering, particularly among highly ranked universities. 

\begin{table*}[t!]
\centering
\begin{tabular}{lcc}
\hline
\textbf{Method} & \textbf{Qwen2.5} & \textbf{GPT-4o-mini} \\
\hline
Non-AHP            & \textbf{24.4\%} & 37.8\% \\
Multi-Agent (Auto) & 20.0\%          & \textbf{42.2\%} \\
\hline
\end{tabular}
\caption{Kendall’s Tau on QS World University Rankings}
\label{tab:qs_ranking}
\end{table*}

\begin{table*}[t!]
\centering
\begin{tabular}{lcc}
\hline
\textbf{Method} & \textbf{Qwen2.5} & \textbf{GPT-4o-mini} \\
\hline
Non-AHP            & 37.8\%          & \textbf{68.9\%} \\
Multi-Agent (Auto) & \textbf{60.0\%} & 60.0\% \\
\hline
\end{tabular}
\caption{Kendall’s Tau on U.S. News Best Global Universities Rankings. 
}
\label{tab:us_ranking}
\end{table*}

\begin{table*}[t]
\centering
\setlength{\tabcolsep}{3pt}
\scriptsize
\begin{tabular}{lcccccccccc}
\toprule
\textbf{Ranking}
& 1 & 2 & 3 & 4 & 5 & 6 & 7 & 8 & 9 & 10 \\
\midrule
QS Ranking
& MIT & Imperial & Stanford & Oxford & Harvard & Cambridge & ETH & NUS & UCL & Caltech \\

\midrule
Zero-Shot (4o-mini)
& MIT & Stanford & Harvard & Caltech & Cambridge & Oxford & ETH & Imperial & UCL & NUS \\

Auto Multi-Agent(4o-mini)
& MIT & Cambridge & Stanford & Harvard & Oxford & Imperial & Caltech & NUS & UCL & ETH \\

\midrule
Zero-Shot (Qwen-2.5)
& Harvard & MIT & Cambridge & Stanford & Caltech & Oxford & Imperial & UCL & NUS & ETH \\

Auto Multi-Agent(Qwen-2.5)
& MIT & Harvard & Caltech & Cambridge & Stanford & Oxford & NUS & Imperial & UCL & ETH \\
\bottomrule
\end{tabular}
\caption{QS Ranking Comparsion }
\label{tab:QS_rank_by_position}
\end{table*}

\begin{table*}[t]
\centering
\setlength{\tabcolsep}{3pt}
\scriptsize
\begin{tabular}{lcccccccccc}

\toprule
\textbf{Ranking}
& 1 & 2 & 3 & 4 & 5 & 6 & 7 & 8 & 9 & 10 \\
\midrule
USNEWS Ranking
& Harvard & MIT & Stanford & Oxford & Cambridge & UC Berkeley & UCL & Washington & Yale & Columbia \\

\midrule
Zero-Shot (4o-mini)
& MIT & Harvard & Stanford & Cambridge & Oxford & Yale & UCL & UC Berkeley & Columbia & Washington \\

Auto Multi-Agent(4o-mini)
& MIT & Harvard & Stanford & Yale & Cambridge & Oxford & UCL & UC Berkeley & Columbia & Washington \\

\midrule
Zero-Shot (Qwen-2.5)
& Cambridge & Oxford & Harvard & MIT & Stanford & Yale & Columbia & UC Berkeley & Washington & UCL \\

Auto Multi-Agent(Qwen-2.5)
& MIT & Harvard & Stanford & Yale & UC Berkeley & Oxford & UCL & Washington & Columbia & Cambridge \\

\bottomrule
\end{tabular}
\caption{QS Ranking Comparsion}
\label{tab:usnews_rank_by_position}
\end{table*}

\section{Prompts}

\label{sec:prompt}


Tables~\ref{ahpInstruction},~\ref{singleagent}, and~\ref{agentPrompts} present the prompt templates used in our experiments.  
All prompts are constructed within the Analytic Hierarchy Process (AHP) framework, which structures decision-making into interpretable steps of problem definition, criteria formulation, pairwise comparison, and final aggregation.

\paragraph{AHP-Instruction.}
The AHP-Instruction prompt, shown in Table~\ref{ahpInstruction}, defines a structured decision-making process based on the AHP.  
It guides the model to define the decision problem, identify 3–5 key criteria, perform pairwise comparisons using the AHP 1–9 scale with reciprocals for less important factors, and generate an interpretable ranking of alternatives.

\paragraph{Multi-Agent Setting.}
The multi-agent framework realizes the AHP reasoning pipeline by decomposing the decision process into specialized agents. In the manual setting (Table~\ref{agentPrompts}), the workflow is divided into four predefined roles: a \textit{Retrieval/Summary Agent} for contextual analysis, a \textit{Criteria Generation Agent} for identifying evaluation criteria, a \textit{Pairwise Weights Agent} for estimating criterion importance via AHP pairwise comparisons, and a \textit{Decision-Making Agent} for aggregating weighted scores and selecting the final alternative. This explicit division of labor improves modularity and makes intermediate reasoning steps transparent and interpretable.

In contrast, the automatic multi-agent setting (Table~\ref{autoagentPrompts}) does not rely on predefined agent roles. Instead, a coordinating \textit{Captain Agent} autonomously organizes and executes the same AHP reasoning pipeline, including problem formulation, criterion identification, weighting, and aggregation. Despite differences in agent instantiation, both settings follow the same underlying AHP formulation, enabling a controlled comparison between manually designed and automatically coordinated multi-agent reasoning.

\clearpage

\begin{table*}[h]
  \renewcommand{\arraystretch}{1.2}
  \centering
  \small
  \begin{tabular}{p{16cm}} 
  \toprule
  \textbf{\textit{System Prompt}} \\
  You are an expert assistant in AHP decision-making. Follow a step-by-step structure.\\

1. Identify the decision problem clearly.\\
2. Identify 3–5 key criteria that influence the alternative decision.\\
3. Use AHP pairwise comparison (1–9 scale, reciprocals for less important) to weight criteria and score alternatives.\\
4. Output the ranking or prioritization in a clear, interpretable format.\\
\bottomrule
  \end{tabular}
\caption{AHP-Instruction Prompt.}
\label{ahpInstruction}
\end{table*}

\begin{table*}[h]
  \renewcommand{\arraystretch}{1.2}
  \centering
  \small
  \begin{tabular}{p{16cm}} 
  \toprule
  \textbf{\textit{System Prompt}} \\
  You are an expert assistant in AHP decision-making. Follow a step-by-step structure.\\
  \midrule
  \textbf{\textit{Step 1}} \\
    Question and Alternatives: \{question\} \\
    Step 1: Identify 3–5 key criteria that influence the alternative decision.\\
  \midrule
  \textbf{\textit{Step 2}} \\
    Question and Alternatives: \{question\}  \\
    Step 2: Using pairwise comparison, assign weights to the criteria and score the alternatives under each criterion using the AHP 1–9 scale. Use reciprocal values (1/x) for less important items.\\
  \midrule
  \textbf{\textit{Step 3}} \\
    Question and Alternatives: \{question\}  \\
    Step 3: Calculate the final scores for each alternative based on the alternative, question and weighted criteria,
    and select the best alternative. \\
  \bottomrule
  \end{tabular}
\caption{Single-Agent decomposition prompt.}
\label{singleagent}
\end{table*}

\begin{table*}[h]
  \renewcommand{\arraystretch}{1.2} 
  \centering
  \small
  \begin{tabular}{p{16cm}} 
  \toprule
  \textbf{\textit{Retrieval/Summary Agent}} \\
  You are an expert assistant in planning analysis. You will be given a question along with a set of alternatives and examples. Your task is to analyze how each alternative relates to the question.\\
  \midrule
  \textbf{\textit{Criteria Generation Agent}} \\
    You are an expert in AHP decision-making.  \\
    You will be given a structure that includes a question, a set of alternative answers and anaysis.\\
    Your task:\\
        - Analyze the question and the alternatives.\\
        - Generate a concise set of evaluation criteria for AHP (3–5 criteria).\\
        - Briefly define each criterion.\\
  \midrule
  \textbf{\textit{Pairwise Weights Agent}} \\
  You are an expert in AHP decision-making.  \\
  You will be given a question with alternatives and a criteria list. Using AHP, do pairwise comparisons with the Saaty 1–9 scale (use reciprocals for the inverse). \\

  \midrule
  \textbf{\textit{Decision Making Agent}} \\
   You are an expert in AHP decision-making.  \\
   You will be given a question with alternatives and a criteria weight list. Identify the best alternative and provide a brief explanation for your choice.\\
  \bottomrule
  \end{tabular}
\caption{Manual multi-agent prompt with predefined agent roles.}
\label{agentPrompts}
\end{table*}

\begin{table*}[h]
  \renewcommand{\arraystretch}{1.2} 
  \centering
  \small
  \begin{tabular}{p{16cm}} 
  \toprule
  \textbf{\textit{Captain Agent}} \\
  You are an expert assistant in AHP decision-making. \\

Follow a step-by-step structure:\\
1. Identify the decision problem clearly.\\
2. Identify 3–5 key criteria that influence the alternative decision.\\
3. Use AHP pairwise comparison (1–9 scale, reciprocals for less important) to weight criteria and score alternatives.\\
4. Output the ranking or prioritization in a clear, interpretable format.\\
  \bottomrule
  \end{tabular}
\caption{Automatic Multi-Agent prompt (Captain Agent).}
\label{autoagentPrompts}
\end{table*}

\end{document}